\documentclass[journal]{IEEEtran}
\usepackage{times}
\usepackage{graphicx}
\usepackage{amsmath}
\usepackage{amssymb}
\usepackage{newunicodechar}
\usepackage{colortbl}
\definecolor{mycyan}{rgb}{.1,0.39,0.66}
\definecolor{myred}{rgb}{.9,0.36,0.36}
\usepackage[breaklinks=true,bookmarks=true,colorlinks, linkcolor=myred, citecolor=mycyan]{hyperref} 
\usepackage{authblk}
\usepackage{multirow}
\usepackage{makecell}
\usepackage[numbers, sort]{natbib}

\usepackage{scrextend}
\deffootnote[0em]{0em}{0em}{\thefootnotemark.\space}

\ifCLASSINFOpdf
\else
\fi
\begin{document}
%
\title{Complementary Feature Enhanced Network with Vision Transformer for Image Dehazing}
%
%
%

\author{Dong Zhao,
        Jia Li, ~\IEEEmembership{Senior Member,~IEEE,}
        Hongyu Li,
        Long Xu
\thanks{D. Zhao, J. Li, and H. Li are with the State Key Laboratory of Virtual Reality Technology and Systems, School of Computer Science and Engineering, Beihang University.}
\thanks{J. Li and L. Xu are with the Key Laboratory of Solar Activity, National Astronomical Observatories, Chinese Academy of Sciences.}
\thanks{J. Li and L. Xu are also with the Peng Cheng Laboratory, Shenzhen 518000, China (e-mail: jiali@buaa.edu.cn)}
\thanks{J. Li is the corresponding author. URL: http://cvteam.net}
}

%
%

\markboth{Journal of  XXXXX,~Vol.~X, No.~X, XX~20XX}%
{Zhao \MakeLowercase{\textit{et al.}}: Complementary Feature Enhanced Network with Vision Transformer for Image Dehazing}
%



\maketitle

\begin{abstract}
~Conventional CNNs-based dehazing models suffer from two essential issues: the dehazing framework (limited in interpretability) and the convolution layers (content-independent and ineffective to learn long-range dependency information). In this paper, firstly, we propose a new complementary feature enhanced framework, in which the complementary features are learned by several complementary subtasks and then together serve to boost the performance of the primary task. One of the prominent advantages of the new framework is that the purposively chosen complementary tasks can focus on learning weakly dependent complementary features, avoiding repetitive and ineffective learning of the networks. We design a new dehazing network based on such a framework. Specifically, we select the intrinsic image decomposition as the complementary tasks, where the reflectance and shading prediction subtasks are used to extract the color-wise and texture-wise complementary features.
To effectively aggregate these complementary features, we propose a complementary features selection module (CFSM) to select the more useful features for image dehazing.
Furthermore, we introduce a new version of vision transformer block, named Hybrid Local-Global Vision Transformer (HyLoG-ViT), and incorporate it within our dehazing networks. The HyLoG-ViT block consists of the local and the global vision transformer paths used to capture local and global dependencies. As a result, the HyLoG-ViT introduces locality in the networks and captures the global and long-range dependencies.
Extensive experiments on homogeneous, non-homogeneous, and nighttime dehazing tasks reveal that the proposed dehazing network can achieve comparable or even better performance than CNNs-based dehazing models.
\end{abstract}

%
\IEEEpeerreviewmaketitle

%
%
%
%

\section{Introduction}
\IEEEPARstart{I}{n} bad weather conditions (such as haze, mist, and fog), the captured outdoor images are usually degraded by the small particles or water droplets suspended in the atmosphere \cite{nayar2002vision}.
Due to the atmospheric scattering, emission and absorption, the degraded images suffer from \emph{color} distortion and \emph{texture} blurring \cite{dong2020physics}. Mathematically, the Atmospheric Scattering Model (ASM) popularly used to describe the hazy images is:
\begin{equation}\label{eq.asm}
{I}_D({p})=({I}_H({p})-{A})/t({p})+{A},
\end{equation}
where ${I}_H$ is the hazy image,  ${I}_D$ is the scene radiance (haze-free image), ${A}$ is the atmospheric light, $t$ is the transmission map, and ${p}$ is the pixel position.

To solve the ill-posed problem of the ASM (\ref{eq.asm}), early \emph{prior-based} single image dehazing methods use handcrafted priors that are observed from the \emph{color} information, such as dark channel prior (DCP) \cite{he2011single}, color attenuation prior (CAP) \cite{zhu2015fast} and haze-line prior (HLP) \cite{berman2018single}, and the \emph{texture} information, such as change of detail prior (CoDP) \cite{coe2015} and gradient channel prior (GCP) \cite{Kaur2020}. However, when the priors are invalid in some instances, these methods may generate unnatural artifacts.

Recently, it has witnessed the flourishing and rapid development of Convolutional Neural Networks (CNNs), and many \emph{CNNs-based} image dehazing methods \cite{cai2016dehazenet,ren2016single,li2017aod} have been proposed to estimate the haze effects. These methods commonly outperform the prior-based method since the deep networks can implicitly learn the rich haze-relevant features and overcome the limitations of a single specific prior \cite{li2020deep}. However, existing CNNs-based dehazing models suffer from essential issues that stem from two aspects: the dehazing framework and the convolution layers.

\begin{figure}[t]
\centering
\includegraphics[width=1\columnwidth]{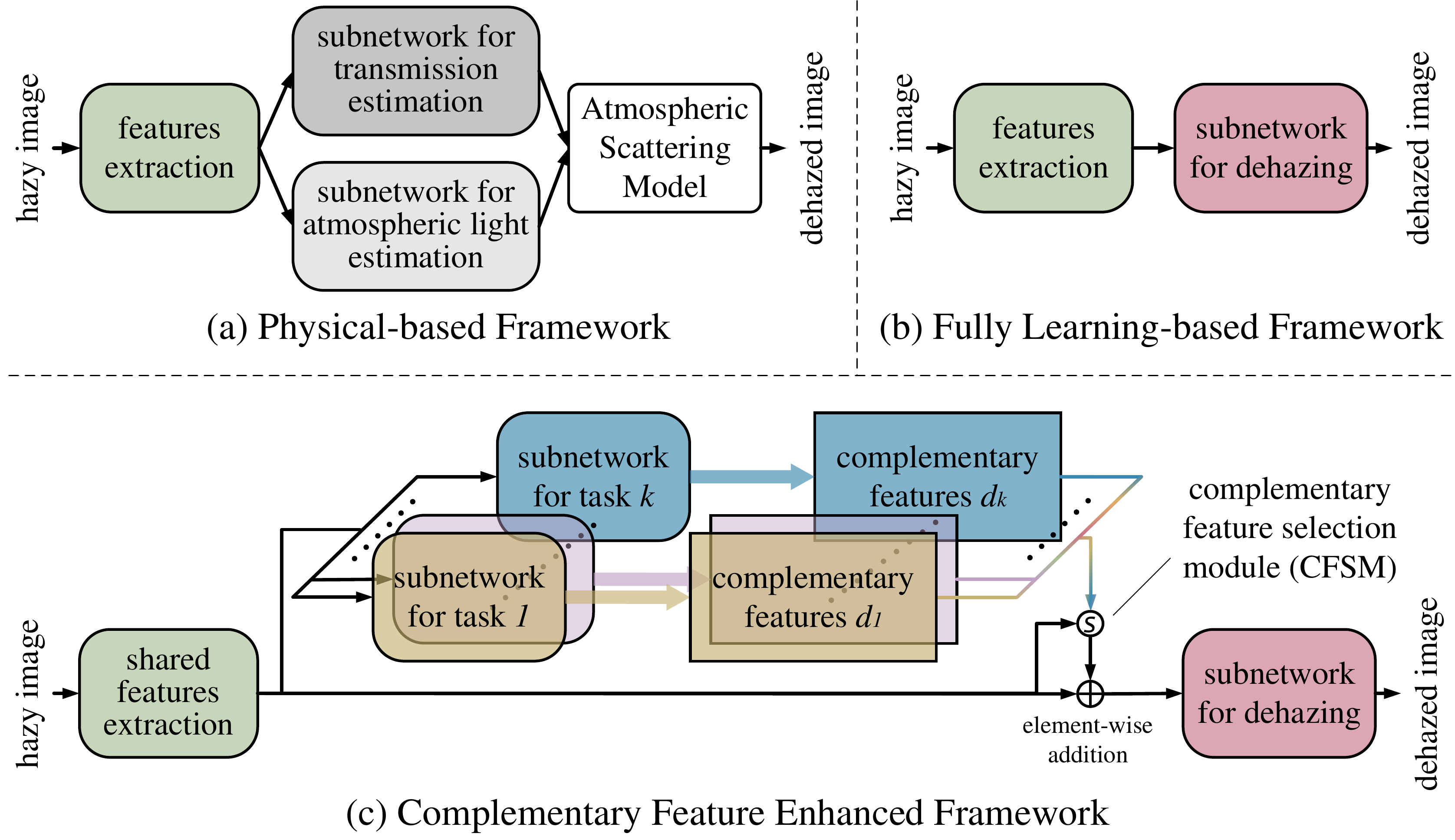} 
\caption{Comparisons with the different CNNs-based dehazing frameworks: (a) physical-based framework, (b) fully learning-based framework, and (c) the proposed complementary feature enhanced framework.}\vspace{-.06in}
\label{fig.frameworks}
\end{figure}

To \emph{the first aspect}, existing CNNs-based dehazing methods can be divided into two categories. The first is the \emph{physical-based} framework (as illustrated in Fig.\ref{fig.frameworks} (a)), such as DehazeNet \cite{cai2016dehazenet}, multi-scale CNN model (MSCNN) \cite{ren2016single} and aggregated transmission propagation networks \cite{Liu2019Learning}, which try to predict the atmospheric light ${A}$ and/or transmission map $t$ at the first step, and then use them to calculate the haze-free image ${I}_D$ according to the ASM (\ref{eq.asm}). However, the ASM cannot completely represent the complex hazy imaging process since it ignores the emission and absorption of atmospheric particles, leading to ineffective dehazing results \cite{li2021dehazeflow}.
Another category is the \emph{fully learning-based} \cite{li2020deep,Chen2018HazeRemoval} (as illustrated in Fig.\ref{fig.frameworks} (b)), including enhanced Pix2pixHD dehazing network (EPDN) \cite{qu2019enhanced}, GridDehazeNet (GDNet) \cite{liu2019griddehazenet}, multi-scale boosted dehazing network (MSBDN) \cite{dong2020multi} and feature fusion attention network (FFA) \cite{qin2020ffa}. Trained in an end-to-end fashion, these models directly recover the haze-free image. However, they have limitations in interpretability \cite{li2020deep} and usually appear ineffective in dense haze removal.
As to \emph{the second aspect}, on the one hand, convolution is content-independent as the same convolution kernel is shared to the entire image, ignoring the distinct nature between different image regions \cite{naseer2021intriguing}. On the other hand, due to inductive biases such as locality, the convolutions are ineffective to learn long-range dependency information \cite{zheng2021rethinking, liang2021swinir}.
The above discussions inspire us to provide a new framework and a more powerful mechanism to replace some of the convolution layers.

To these ends, we first introduce a novel complementary feature enhanced framework, as illustrated in Fig.\ref{fig.frameworks} (c). Unlike previous frameworks that learn the numbers of various features at one time implicitly and inefficiently, the core idea behind the new framework is: different complementary subtasks focus on learning different specific features, i.e. the task-relevant complementary features, respectively; then, these complementary features are aggregated together and served to the primary task, i.e. the dehazing.
We design a new dehazing network based on such framework, where specific complementary features are explored by corresponding subnetworks and then collaborated within the dehazing network.
Motivated by the observations that \emph{color} and \emph{texture} perceptions are the critical visual cues toward identifying objects and understanding scenes \cite{Texture,julesz1962visual}, we select the intrinsic image decomposition as the `complementary tasks', in which the intermediate features learned from reflectance prediction are served as the `color-wise' complementary features, and the `texture-wise' ones are learned from the shading prediction.
That is, our method jointly learns intrinsic image decomposition and haze removal from a single image.
Notice that directly aggregating the redundant, complementary features is inefficient, as the complementary subnetworks may output some haze-irrelevant features.
Therefore, we further propose a Complementary Features Selection Module (CFSM) to automatically select the `right' features that are helpful to the dehazing task and weaken the irrelevant ones.

Very recently, vision transformer \cite{vit} has been known as an alternative to CNNs to learn visual representations due to its content-dependent interactions \cite{vaswani2021scaling} and flexibility in modelling long-range dependencies \cite{chu2021twins}. However, the quadratic increase in complexity with image size hinders its application on dehazing tasks, which requires high-resolution feature maps. Additionally, it is undeniable that global and local information aggregation are useful for low-level vision tasks, while the vision transformer does not possess the locality \cite{vit}. Inspired by these researches, we propose a Hybrid Local-Global Vision Transformer (HyLoG-ViT), which consists of the local and global vision transformer paths. In the local vision transformer path, the standard transformer blocks are operated in a grid of non-overlapped regions, enabling the model to capture the fine-grained and short-distance information within the local regions. In the global vision transformer path, one transformer block is operated on the downscaled feature maps to capture the global and long-range dependencies.
Then, the features from the two paths are hybridized by a convolution layer to improve the local continuity.
Compared with the vanilla vision transformer architecture, the HyLoG-ViT has lower computational complexity and brings locality mechanisms to the networks.

Incorporating the HyLoG-ViT within the complementary feature enhanced framework, we build our dehazing network. Previous works designed specific approaches for different hazy scenes, such as the two-branch neural network (TBNN) \cite{twobranch2021cvprw}, discrete wavelet transform GAN (DW-GAN) \cite{dwgan2021cvprw} and ensemble dehazing networks (EDN) \cite{edn2020cvprw} for non-homogeneous dehazing, and maximum reflectance prior (MRP) \cite{zhang2017fast} and optimal-scale fusion-based dehazing (OSFD) \cite{zhang2020nighttime} methods for nighttime dehazing. Interestingly, extensive experiments on homogeneous, non-homogeneous, and nighttime dehazing tasks reveal that our network exhibits good generalization performances on different hazy scenes without any changes in the network architecture during the training.
The main contributions of this work are summarized as follows:
\begin{itemize}
\item We propose a new framework and built a complementary feature enhanced network for image dehazing by jointly learning the intrinsic image decomposition and image dehazing. The reflectance and shading prediction tasks provide rich complementary features for the dehazing task, enabling the network to generate high-quality haze-free images with natural color and fine details.
\item To effectively fuse the complementary features, we propose a Complementary Features Selection Module (CFSM). The CFSM considerably improves the effectiveness of feature aggregation by adaptively enhancing the proper complementary feature channels while weakening the irrelevant ones.
\item We propose a new variant of vision transformer, namely Hybrid Local-Global Vision Transformer (HyLoG-ViT), which can model both local and global dependencies with lower computational cost than the vanilla vision transformer. With the HyLoG-ViT,  we propose a transformer-based dehazing network for the first time.
\end{itemize}

\begin{figure*}[t!]
\centering
\includegraphics[width=1\textwidth]{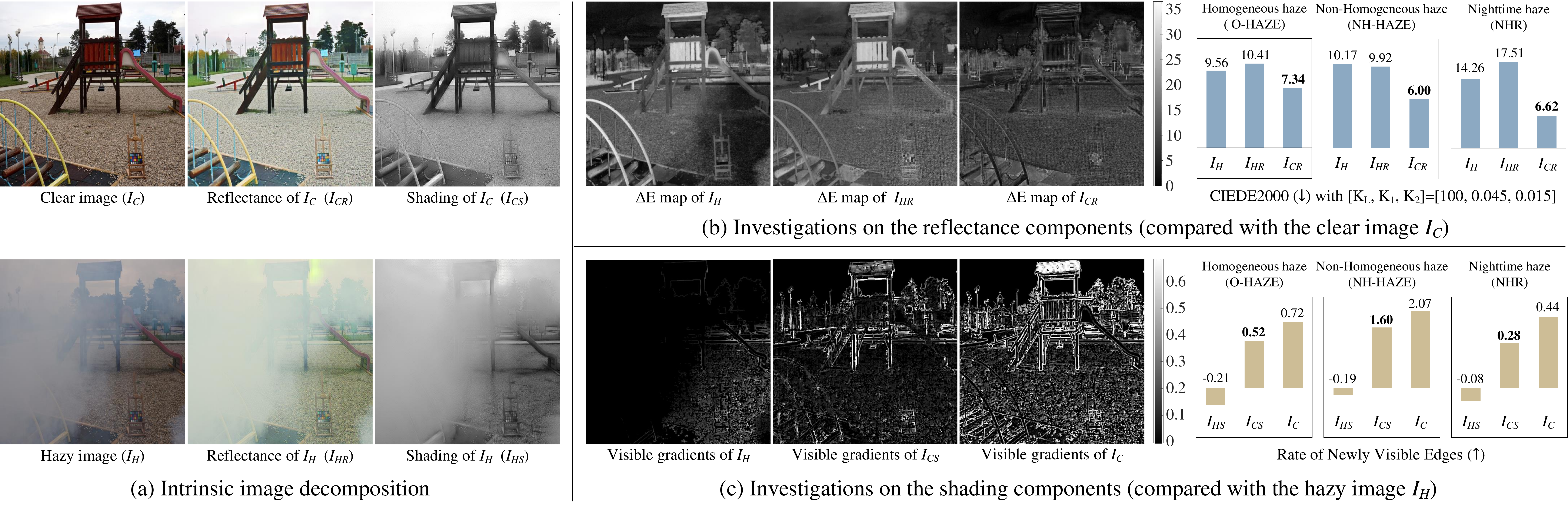}
\caption{Investigations on the intrinsic image decomposition. (a) The reflectance and shading maps of clear and hazy images. (b) Investigations on the reflectance, including $\Delta E$ maps (left) and CIEDE2000 (right) scores of hazy image $I_H$, hazy reflectance $I_{HR}$ and clear reflectance $I_{CR}$ compared with the clear image $I_C$. The $I_{CR}$ gets the lowest scores on both of the three datasets, indicating that $I_{CR}$ preserves similar color information as the $I_{C}$. (c) Investigations on the shading, including the visible gradients maps (left) and the rate of newly visible edges indicators (right) of the hazy shading $I_{HS}$, clear shading $I_{CS}$ and clear image $I_C$. As we can find, $I_{CS}$ and $I_{C}$ contains rich edges information with higher scores. }\vspace{-.06in}
\label{fig.rs}
\end{figure*}

\section{Related Work}
\subsection{Single Image Dehazing}
Single image dehazing is an ill-posed problem. Without any auxiliary information, previous dehazing algorithms require the introduction of specific prior knowledge. Through research in the last decade, many priors have been proposed based on the cues of color (such as the DCP \cite{he2011single}, CAP \cite{zhu2015fast}, and HLP \cite{berman2018single}) and texture (such as the CoDP \cite{coe2015} and GCP \cite{Kaur2020}). For example, Berman et al. \cite{berman2018single} find that for a haze-free image, its pixels that belong to the same color cluster form to a point while forming a line in a hazy image, namely the haze-line. The GCP uses the image gradient to estimate the depth information and the atmospheric light, preserving texture details more efficiently.

Recently, CNNs-based dehazing networks have been extensively studied. One kind of CNNs-based network restores the clear image by estimating the intermediate variables in the atmospheric scattering model \cite{narasimhan2002vision} and then calculates the clear image. For example, the multi-scale CNNs dehazing model (MSCNN) \cite{ren2016single} utilizes a coarse-scale network to estimate the complete transmission map and use a fine-scale network to refine the results.
More recent CNNs-based networks tend to learn hazy-to-clear image translation directly \cite{wu2021contrastive}.
For example, the MSBDN \cite{dong2020multi} removes haze in an end-to-end manner by incorporating boosting and error feedback principles into a U-Net \cite{ronneberger2015u} architecture with a dense feature fusion module. 

Unlike the previous dehazing frameworks, our method relies neither on the atmospheric scattering model nor on a completely black box system. The subnetworks of intrinsic image decomposition provide color- and texture-wise complementary features, enabling the dehazing subnetwork to yield haze-free images with natural color and fine details.

\emph{Non-homogeneous and nighttime dehazing}.
For different complicated haze patterns, the dehazing method should be specifically designed,  such as the TBNN \cite{twobranch2021cvprw}, DW-GAN \cite{dwgan2021cvprw} and EDN \cite{edn2020cvprw} for non-homogeneous dehazing, and MRP \cite{zhang2017fast} and OSFD \cite{zhang2020nighttime} methods for nighttime dehazing. Unlike these models, training on different datasets, our network exhibits good generalization performances on homogeneous, non-homogeneous, and nighttime dehazing tasks without any changes in the network architecture.

\subsection{Vision Transformer}
Very recently, vision transformers \cite{vit} have received increasing research interest in image and video vision tasks, including object detection \cite{dert}, image classification \cite{vit} and semantic segmentation \cite{zheng2021rethinking}.
Many new versions of vision transformers \cite{pvt,liu2021swin,rao2021dynamicvit,Chen_2021_ICCV} have been proposed to relieve the high computational cost problems. For example, the Swin Transformer \cite{liu2021swin} uses a locally-grouped self-attention \cite{chu2021twins}, where the input features are separated into a grid of non-overlapped windows and the vision transformer is operated only within each window. 
Many other methods have been proposed to bring inductive biases into the vision transformer \cite{wu2021cvt,li2021localvit,chu2021twins,yang2021focal}, such as the LocalViT \cite{li2021localvit}, which brings a locality mechanism to ViT by employing the depth-wise convolution into the feed-forward network. CvT \cite{wu2021cvt} proposes a convolutional token embedding to model local spatial contexts and a convolutional projection layer to provide efficiency benefits.

New versions of vision transformers are also developed and used in low-level vision tasks. For example, the Uformer \cite{wang2021uformer} designs a general U-shaped Transformer-based structure for image restoration. It also proposes a locally-enhanced window Transformer block to reduce the computation cost. The SwinIR \cite{liang2021swinir} utilizes residual Swin Transformer blocks to extract deep features for image restoration.

In our HyLoG-ViT, the local vision transformer path is similar to the LeWin transformer proposed in Uformer. However, the local version used in Uformer alone fails to effectively model global dependencies and preserve the local continuity around those regions. By contrast, our HyLoG-ViT block can capture both local and global dependencies simultaneously.


\section{Method}

\begin{figure*}[ht!]
\centering
\includegraphics[width=.9\textwidth]{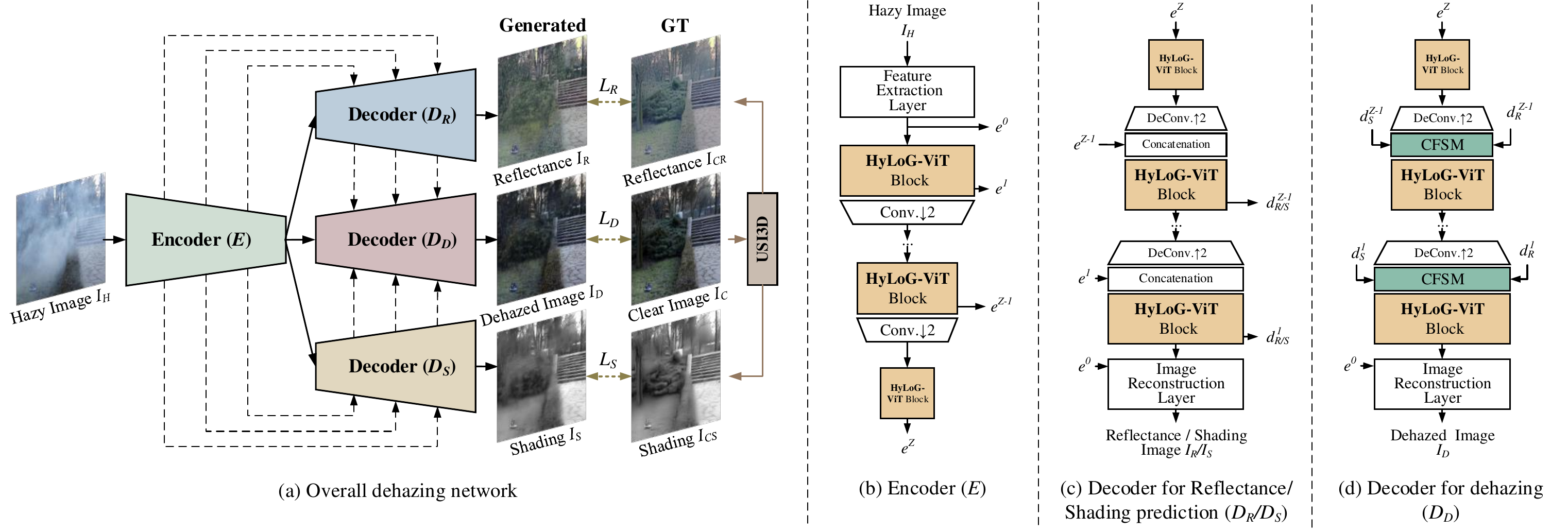}
\caption{Architectures of our dehazing network. (a): the overall dehazing network, which consists of an encoder and three different decoders. (b): the architecture of the shared encoder $E$. (c): the architecture of decoder $D_R$/$D_S$. (d): the architecture of decoder $D_D$. USI3D: a pre-trained unsupervised intrinsic image decomposition model proposed by \cite{Liu2020Unsupervised}. CFSM: the complementary features selection module. $L_R$, $L_S$ and $L_D$: loss functions for the three tasks.}\vspace{-.06in}
\label{fig.network}
\end{figure*}
\subsection{Complementary Features Enhanced Framework}


To achieve high performance of haze removal, we propose a novel framework from a new perspective, namely the complementary feature enhanced framework. In this framework, each group of \emph{complementary features} are learned by corresponding complementary tasks, and together they serve the primary task (i.e. the dehazing task in this paper). Unlike previous CNNs-based approaches that learn haze-relevant features implicitly, complementary features here can be purposively chosen and learned by training under specific subnetworks and datasets. Problems that arise here are: \emph{which complementary features} should be considered for image dehazing, and \emph{which complementary tasks} are used to learn such features effectively?

Before answering these problems, we here analyse the essential criterion the complementary features should follow. First, each kind of complementary feature represents one interference factor closely related to the prime task. Second, the different complementary features should be weakly correlated with each other so that each complement subtask can focus on doing one thing more effectively. Otherwise, multiple subnetworks learn similar representations, making the network very inefficient. That is, `Each performs one's own functions; together serve their common mission'.

\emph{Complementary Features:} Intuitively, the images captured in hazy or foggy scenes suffer from some interference factors, and the most conspicuous ones are the variation of color and blurriness of edges. Many previous works have witnessed that the information of color \cite{he2011single,zhu2015fast,berman2018single} and texture \cite{coe2015,Kaur2020} are the crucial priors for estimating the haze distribution. Inspired by these observations, we endeavour to build the corresponding subnetworks to learn the color-wise and texture-wise complementary features.

\emph{Complementary Tasks:} In further investigations, we fund that the intrinsic image decomposition (which decomposes an image into reflectance and shading, as shown in Fig.\ref{fig.rs} (a)) is a good candidate as the complementary task; as the reflectance component contains the actual color of the scenes, and the shading component contains the structure and texture information \cite{baslamisli2018joint}. To demonstrate these, we analyse the characters of the reflectance and shading of the hazy and clear images on the homogeneous, non-homogeneous and nighttime haze datasets. Fig.\ref{fig.rs} (b) shows the $\Delta E$ maps and CIEDE2000 \cite{sharma2005ciede2000}\footnote{\noindent We set the Luminance coefficient $K_L$, weighting factors $K_1$ and $K_2$ here are 100, 0.045 and 0.015, respectively, to offset the impact of the luminance component.} metrics of the hazy image ${I}_H$, hazy reflectance ${I}_{HR}$ and clear reflectance ${I}_{CR}$ compared with the clear image ${I}_C$. The investigations demonstrate that the clear reflectance ${I}_{CR}$ maintain the color information of clear image ${I}_C$ with very low $\Delta E$ and CIEDE2000. Similarly, we visualise the visible gradients maps and calculate the Rate of Newly Visible Edges $r_{nve}$ \cite{hautiere2008blind} of hazy shading ${I}_{HS}$, clear shading ${I}_{CS}$ and clear image ${I}_{C}$ compared with the hazy image ${I}_C$, as illustrated in Fig.\ref{fig.rs} (c). We find that the $r_{nve}$ of ${I}_{CS}$ is very close to the one of ${I}_{C}$ on the tree datasets, indicating that the shading ${I}_{CS}$ preserves rich edge and texture information.

Finally, we choose the Unsupervised Learning for Intrinsic Image Decomposition (USI3D) \cite{Liu2020Unsupervised} method to generate clear reflectance and shading samples in our training datasets. Most importantly, the USI3D assumes reflectance-shading independence, coinciding with the second criterion of complementary features as aforementioned. The reflectance predilection subnetwork in our model can focus on color-wise complementary features learning, avoiding the interferences of shading components; and vice versa.




\subsection{Dehazing Network}
In our dehazing model, the intrinsic image decomposition and dehazing are considered a joint model by exploring the former's complementary features that the latter requires. Specifically, our network is a multi-task learning model with a decoder-focused architecture \cite{vandenhende2021multi}.

\subsubsection{Overall Architecture}
The overview of our dehazing network consists of a shared encoder and three decoders, as illustrated in Fig.\ref{fig.network}. Denote the hazy image is ${I}_H$, the shared encoder $E$ is used to extract the shallow and deep features:
\begin{equation}\label{eq.encoder}
{e}^0= F_{E}^0({I}_H), ~{e}^z= F_{E}^z({e}^{z-1}),
\end{equation}
where $F_{E}^0$ denotes the feature extraction layer used to extract shallow feature ${e}^0$; $F_{E}^z$ denotes $z$-th stage of encoder $E$, ${e}^z$ refers to the deep feature at stage $z$. $z\in[1, \cdots, Z]$, and $Z$ is the total number of stages.
Three parallel decoders follow the encoder. Decoders $D_R$ and $D_S$ are used to predict the reflectance and shading of the haze-free image, respectively, and their intermediate features are served as complementary features to the decoder $D_D$ to generate the high-quality haze-free image. The decoders are described as:
\begin{equation}\label{eq.decoders_r}
{{d}}_{{R}}^Z = F_{D_R}^Z({{d}}_E^Z), ~{{d}}_{{R}}^z = F_{D_R}^z({d}_{R}^{z-1}, {e}^z),
\end{equation}
\begin{equation}\label{eq.decoders_s}
{{d}}_{{S}}^Z = F_{D_S}^Z({{d}}_E^Z), ~{{d}}_{{S}}^z = F_{D_S}^z( {d}_{S}^{z-1}, {e}^z),
\end{equation}
\begin{equation}\label{eq.decoders_d}
{{d}}_{{D}}^Z=F_{D_D}^Z({e}^Z, {d}_{R}^Z, {d}_{S}^Z), ~{{d}}_{{D}}^z=F_{D_D}^z({d}_{D}^{z-1}, {d}_{R}^z, {d}_{S}^z),
\end{equation}
where  ${F}_{D_{R}}^z$, ${F}_{D_{S}}^z$ and ${F}_{D_{D}}^z$ denote the $z$-th stage of decoder $D_R$, $D_S$ and $D_D$, respectively, $z\in[1, \cdots, Z]$.
${d}_{{R}}^z$, ${d}_{{R}}^z$ and ${{d}}_{{R}}^z$ are the intermediate features of the decoder $D_R$, $D_S$ and $D_D$ at stage $z$, respectively.
The final reflectance ${I}_R$, shading $I_S$, and haze-free image ${I}_D$ are generated through three image reconstruction layers, i.e., $F_{D_R}^0$, $F_{D_S}^0$ and $F_{D_D}^0$, respectively:
\begin{equation}\label{eq.ir}
 \centering
 \begin{aligned}
{I}_R =& F_{D_R}^0({d}_{R}^{1}, {e}^0),~~~I_S = F_{D_S}^0({d}_{S}^{1}, {e}^0),\\
{I}_D=&F_{D_D}^0({d}_{D}^1, {d}_{R}^1, {d}_{S}^1, {e}^0).
 \end{aligned}
\end{equation}

\subsubsection{Details of the Encoder and Decoder}
The encoder $E$ comprises of a feature extraction layer and a series of HyLoG-ViT blocks (see section \ref{ss.hylog}). The extraction layer is built by a $5\times5$ convolution and a basic ResNet block \cite{he2016deep} to extract shallow features. Each HyLoG-ViT block is followed by a 4$\times$4 convolution to downscale the spatial resolution with stride 2 and double the channel number. The convolutions used here can also bring the inductive bias into this `transformer-based' encoder.
In the decoder $D_R$ and $D_S$, except the bottom stage $Z$, feature $d_{R/S}^{z-1}$ from the previous decoder block is first upscaled by a 4$\times$4 deconvolution to expand the spatial resolution with stride 2 and halve the channel number. Then, the output is concatenated with the feature $e^{z}$ from the same stage of encoder $E$. Therefore, the subnetworks $E$-$D_R$ and $E$-$D_S$ are formed into two U-shaped structures, which alleviates the issue of spatial information loss caused by downscaling.
Different from decoder $D_R$ and $D_S$, for $z$-th stage in $D_D$, there have three inputs: the feature $d_D^{z-1}$ from the previous stage; the complementary features $d_R^z$ and $d_S^z$ from the same stage of $D_R$ and $D_S$, respectively. These input features are fed into a complementary features selection module to select the most useful channels from $d_R^z$ and $d_S^z$ dynamically.

\subsection{Complementary Features Selection Module (CFSM)}\label{ss.cfsm}
\begin{figure}[t]
\centering
\includegraphics[width=.9\columnwidth]{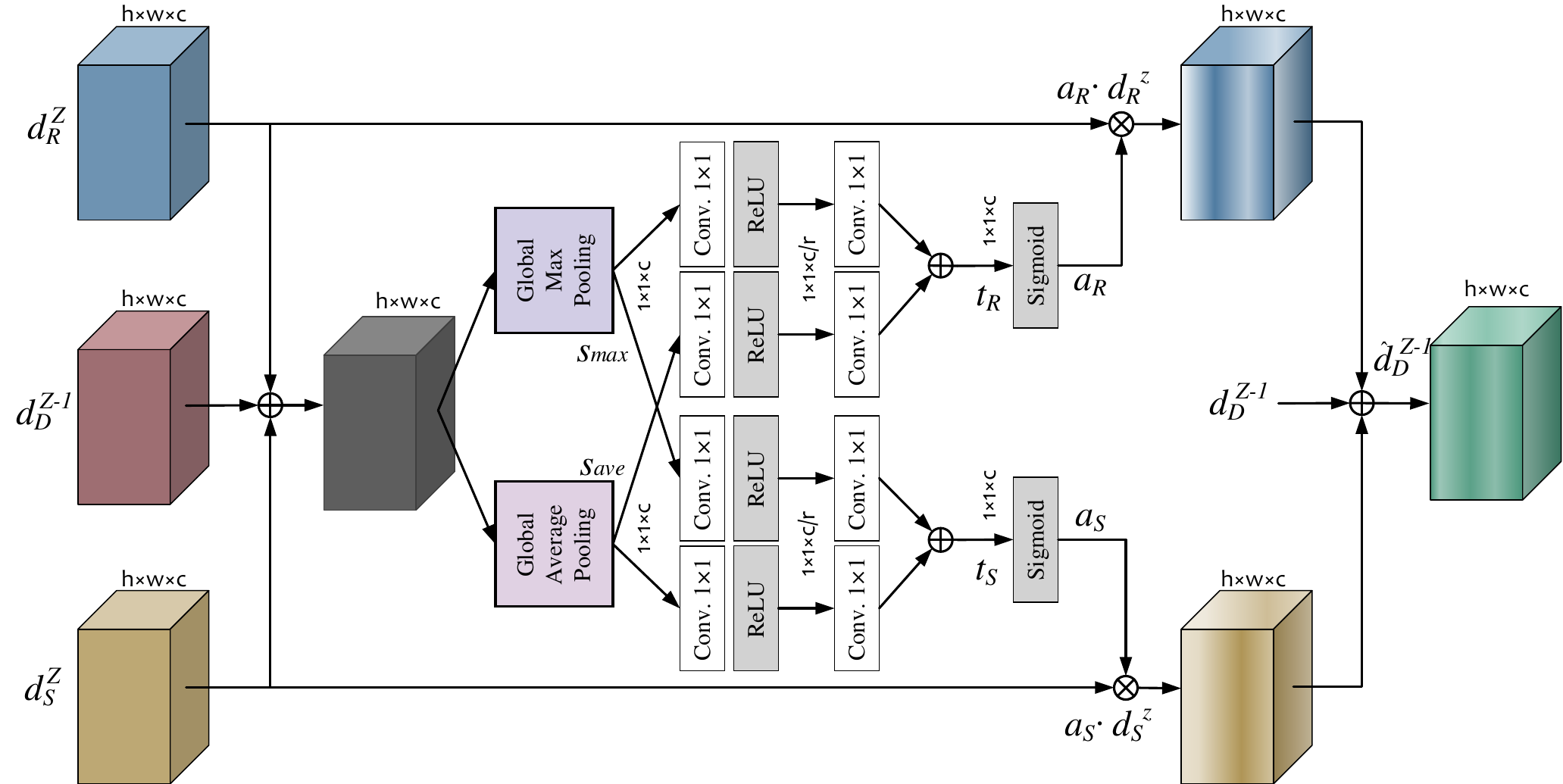}
\caption{Schematic for the complementary features selection module (CFSM). $\oplus$ denotes the element-wise summation. $\otimes$ denotes the element-wise production.}\vspace{-.06in}
\label{fig.cfsm}
\end{figure}

One can simply aggregate the complementary features $d_R^z$ and $d_S^z$ via element-wise summation or concatenation operation, which, however, is inefficient. Therefore, we propose the CFSM to fuse the complementary features in a nonlinear fashion. The architecture of the CFSM is illustrated in Fig.\ref{fig.cfsm}.

Given the complementary features $d_R^z$ and $d_S^z$ and the feature $d_D^{z-1}$ ($\in\mathbb{R}^{h\times w\times c}$), they are first combined via element-wise summation. Then, the outputs are transformed to two channel-wise \emph{statistics} $s_{ave}$ and $s_{max}$ ($\in \mathbb{R}^{1\times 1\times c}$) by a global average pooling and a global max pooling, respectively. Each statistic is separated into two streams: one stream for $d_R^z$ feature selection and another for $d_S^z$. Take $d_R^z$ stream as an example, the $s_{ave}$ and $s_{max}$ are followed by channel-downscaling $1\times 1$ convolution layers to calculate the \emph{compact feature vectors}, $v_{ave\_R}$ and $v_{max\_R}$ ($\in\mathbb{R}^{1\times 1\times c/r}$, where $r=4$ in our model), respectively. The feature vectors are fed into two parallel channel-upscaling $1\times 1$ convolution layers and provide \emph{feature descriptors} $t_{ave\_R}$ and $t_{max\_R}$ ($\in\mathbb{R}^{1\times 1\times c}$). The final feature descriptor is defined as $t_R=t_{ave\_R}+t_{max\_R}$. The $t_R$ passes through the Sigmoid function to generate \emph{attention score} $a_R$ ($\in\mathbb{R}^{1\times 1\times c}$) for $d_R^z$. Similar, we can get the attention score $a_S$ for $d_S^z$. The overall process of feature recalibration and aggregation is defined as:
\begin{equation}\label{eq.ir_d}
\widehat{d}_D^{z-1}={d}_D^{z-1} + a_R \cdot d_R^z + a_S \cdot d_S^z.
\end{equation}
Note that our CFSM utilizes the global average- and max-pooling to gather important clues about complementary features, increasing the representation power and can inferring finer channel-wise attention \cite{woo2018cbam}.

\subsection{Hybrid Local-Global Vision Transformer (HyLoG-ViT)}\label{ss.hylog}

\begin{figure}[t]
\centering
\includegraphics[width=1\columnwidth]{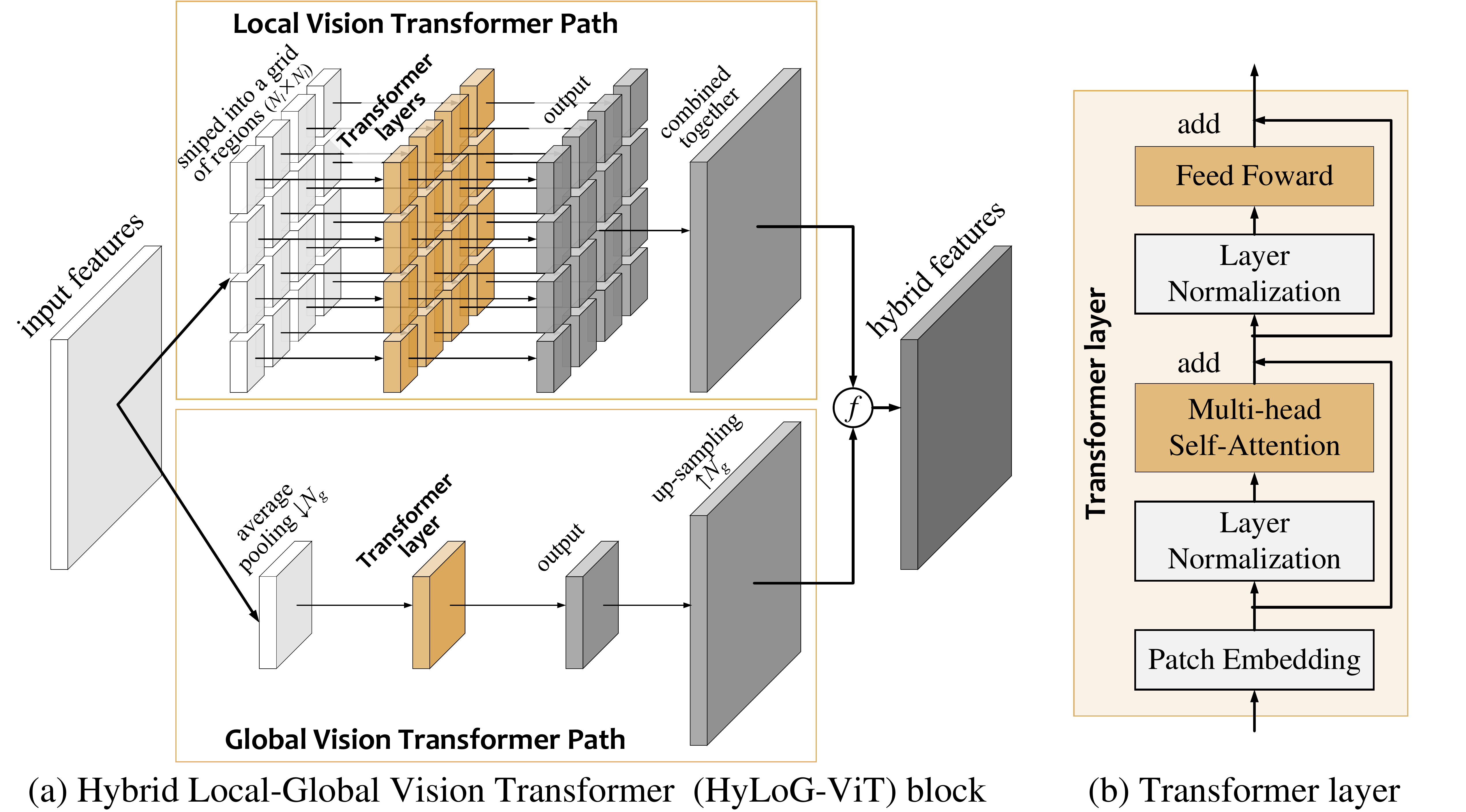} 
\caption{Diagram of the hybrid local-global vision transformer (HyLoG-ViT). $\textcircled{\emph{\footnotesize{f}}}$ represents the fuse operation.}
\label{fig.hlgvit}\vspace{-.06in}
\end{figure}

We propose a HyLoG-ViT model to address the challenges of high computation cost and lack of locality. As shown in Fig.\ref{fig.hlgvit}, a HyLoG-ViT block consists of two paths. In the local vision transformer path, the input feature map is grouped into grids of non-overlapped regions, and the transformer block is applied only within each region.
Given the feature maps $X \in \mathbb{R}^{H\times W\times C}$ where the $H$, $W$ and $C$ are the height, width and channel number of the maps, the computation of the local vision transformer path can be expressed as:
 \begin{equation}\label{eq.lvit}
 \begin{aligned}
X&= \{X^1, X^2, \cdots , X^{ N_l\times N_l}\}; \\
X_l^i&=Transformer(X^i), ~~i\in[1, 2,\cdots, N_l\times N_l];\\
X_l&=\{X_l^1, X_l^2, \cdots , X^{ N_l\times N_l}\},\\
\end{aligned}
\end{equation}
where $X_l^i$ is the output of the $i$-th region, $X_l$ is the combined output of the local vision transformer path; $Transformer(\cdot)$ represents the standard transformer layer, as illustrated in Fig.\ref{fig.hlgvit} (b). $N_l$ is the region number per column/row.
This design enables the vision transformer to focus on capturing region-level attention and exploring local context information.
However, the local vision transformer path cannot model global dependencies and lose local continuity around those regions. Therefore, we also introduce the global vision transformer path, where the input feature is down-sampled by the average pooling operation to reduce the spatial resolution. The output is fed into the standard transformer layer. The computation of the global vision transformer path is:
 \begin{equation}\label{eq.gvit}
 \begin{aligned}
X_g= &Upsamp_{\uparrow N_g}(Transformer(Avepool_{\downarrow N_g}(X)));
\end{aligned}
\end{equation}
where $X_g$ is the output of the global vision transformer path. $Avepool_{\downarrow N_g}(\cdot)$ is the average pooling with reduction ratio $N_g$; $Upsamp_{\uparrow N_g}(\cdot)$ represents the upsampling operation with upscale ratio $N_g$.
The global vision transformer path improves efficiency while still maintaining the capability of aggregating global information.
The hybrid outputs of the two paths are concatenated and transformed to the original dimension by a $3\times 3$ convolution layer:
\begin{equation}\label{eq.hybrid}
 \begin{aligned}
X_h= Conv_{3\times 3}(Concat(X_l, X_g));
\end{aligned}
\end{equation}
where $X_h$ is the final output of HyLoG-ViT block; $Concat(\cdot)$ is the channel-wise concatenation; $Conv_{3\times 3}(\cdot)$ is a $3\times 3$ convolution layer.
 This fuse operation maintains the merits of the local and global vision transformer and introduces locality into the network.

\emph{{Positional Encoding}}:
The positional encoding used in transformers aims to retain positional information. However, for low-level vision tasks, the input images with varying sizes are commonly different from the training ones. As a result, it may decrease performance \cite{chu2021conditional, xie2021segformer} and break the translation invariance \cite{chu2021twins}.
Note that our model can achieve good results even without additional positional encoding modules.  We argue that the $3\times 3$ convolution used in (\ref{eq.hybrid}) is sufficient to provide positional information for Transformers.

\emph{{Complexity Analysis}}:
For the local vision transformer path, denote the input feature $X\in\mathbb{R}^{H\times W \times C}$ is grouped into $N_l\times N_l$ regions and the spatial-resolution of each region is $\frac{H}{N_l}\times \frac{W}{N_l} \times C$. Then, the complexity of the self-attention in the local vision transformer path is reduced from $O((HW)^2 C)$ to $O(\frac{(HW)^2}{N_l^2} C)$ compared with the standard self-attention.
For the global vision transformer path, denote the spatial dimension of the input feature is downscaled by a reduction ratio $N_g$, and the complexity is $O(\frac{(HW)^2}{N_g^2} C)$. Therefore, the total self-attention computational complexity of HyLoG-ViT block is  $O((\frac{1}{N_l^2}+\frac{1}{N_g^2}) (HW)^2C)$.

\subsection{Loss Function}\label{ss.loss}
In our training, the ground truths of reflectance $I_{CR}$ and shading $I_{CS}$ are generated by operating the pre-trained intrinsic image decomposition model USI3D \cite{Liu2020Unsupervised} on the ground truth of clear image $I_{C}$.
For reflectance and haze-free image estimations, we train the networks using the L2 reconstruction loss, conditional adversarial loss \cite{mirza2014conditional,isola2017image}, and SSIM loss (as used in \cite{cai2018learning}).
For the shading estimation, we use the L2 reconstruction loss and an edge preserve loss which is defined as:

 \begin{equation}\label{eq.lossedge}
 \begin{aligned}
{L}_{{e}}=&\mathbb{E}_{{I}_S}[ || \triangledown_x {I}_{S}({p})-\triangledown_x {I}_{CS}({p})||_2 \\
&+ ||\triangledown_y {I}_{S}({p}))- \triangledown_y {I}_{CS}({p}) ||_2],
\end{aligned}
\end{equation}
where $\mathbb{E}$ is the mean operation on a batch of samples; $\triangledown_x$ and $\triangledown_y$ are the spatial derivatives at $x$ and $y$ directions, respectively. $p$ refers to the spatial location.
Overall, the hybrid loss function contains three parts, that is:
\begin{equation}\label{eq.loss_overall}
{L}= \lambda_{R}{L}_{R} + \lambda_{S}{L}_{S} + \lambda_{{D}}{L}_{{D}},
\end{equation}
where ${L}_{R}$, ${L}_{S}$, and ${L}_{D}$ are the loss functions for the three vision tasks, respectively.

\begin{figure*}[t!]
\centering
\includegraphics[width=1\textwidth]{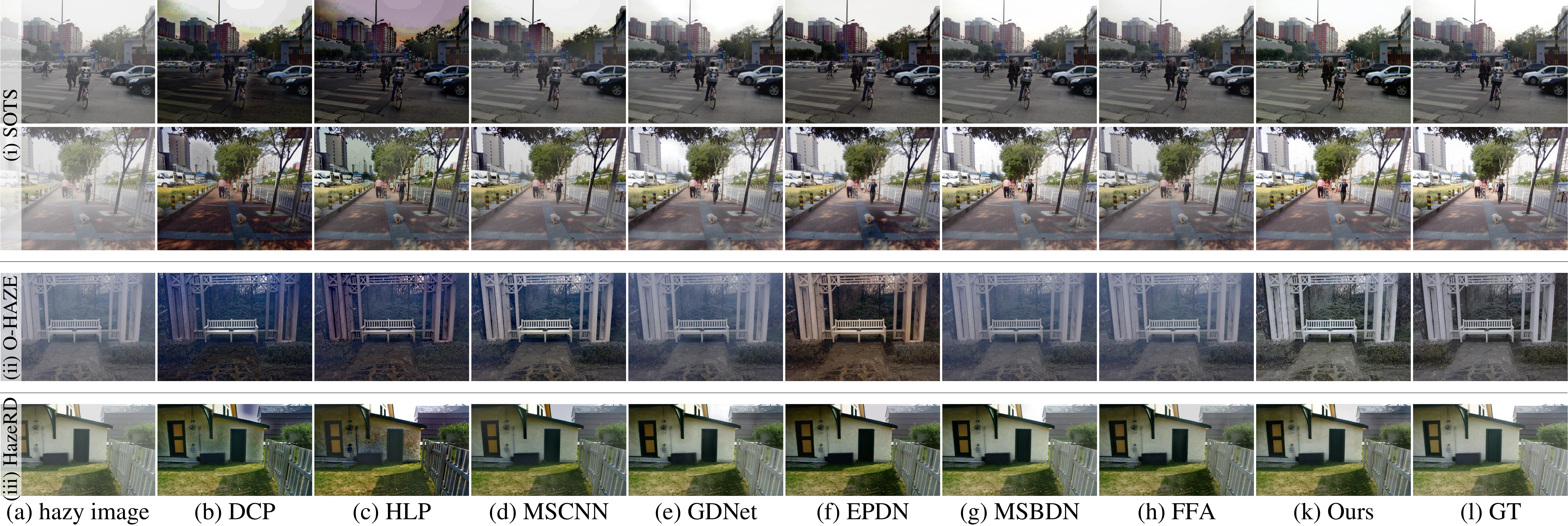}
\caption{Dehazing results of the \emph{Synthetic Homogeneous} hazy images (on the RESIDE, O-HAZE and HazeRD datasets). (a): the hazy images. (b)-(k): the dehazing results of DCP \cite{he2011single}, HLP \cite{berman2018single}, MSCNN \cite{ren2016single}, GDNet \cite{liu2019griddehazenet}, EPDN \cite{qu2019enhanced}, MSBDN \cite{dong2020multi}, FFA \cite{qin2020ffa} and Ours, respectively. (l): the ground truth.} \vspace{-.06in}
\label{fig.result_hd}
\end{figure*}

\begin{figure*}[t!]
\centering
\includegraphics[width=1\textwidth]{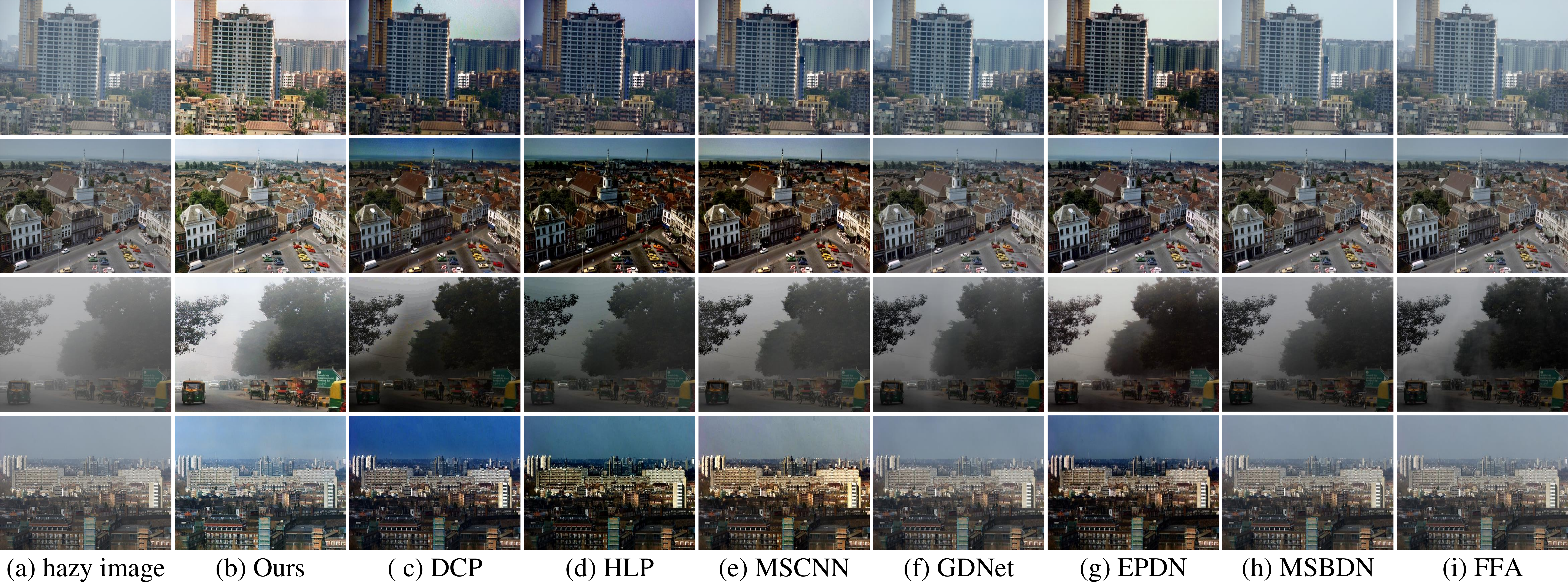}
\caption{Dehazing results of the \emph{Real-World Homogeneous} hazy images. (a): the hazy images. (b): Our dehazing results. (c)-(k): the dehazing results of Ours, DCP \cite{he2011single}, HLP \cite{berman2018single}, MSCNN \cite{ren2016single}, GDNet \cite{liu2019griddehazenet}, EPDN \cite{qu2019enhanced}, MSBDN \cite{dong2020multi} and FFA \cite{qin2020ffa}, respectively.}\vspace{-.06in}
\label{fig.result_hdrw}
\end{figure*}
\begin{table*}[t]
\centering
\caption{Homogeneous image dehazing results on the RESIDE, O-HAZE, HazeRD, and Real-World datasets.$^{\rm \textcolor{myred}{[a]}}$}
\footnotesize
\begin{tabular}{c|r|rrrrrrrrrrrrr}
\hline %
Dataset & Metric & {\textbf{DCP}} & {\textbf{HLP}} & {\textbf{MSCNN}} & {\textbf{GDNet}} & {\textbf{EPDN}} & {\textbf{MSBDN}}  & {\textbf{FFA}}  & {\textbf{Ours}}   \\
\hline
 \multirow{4}*{{\textbf{SOTS}}}&PSNR$\uparrow$ &14.57& 16.44 & 19.15&29.78 &27.42& \textbf{33.79} &29.92 & \underline{32.17} \\
   & SSIM$\uparrow$& 0.749 &0.784& 0.855&\underline{0.978}&0.946&\textbf{0.985}&0.975 &0.970\\
&LPIPS$\downarrow$ &0.157&0.140&0.112&\textbf{0.036}&0.056&\underline{0.038}&0.043& {0.041} \\
   & CIEDE200 $\downarrow$ & 33.01 & 28.65 & 21.65 & \underline{10.86} & 25.55 & \textbf{10.36} & 11.24 & 14.32\\
    \hline
    \hline
  \multirow{4}*{{\textbf{O-HAZE}}}&   PSNR$\uparrow$ & 15.09&15.28&\underline{17.22}&16.67&17.17&16.76&16.09 &\textbf{29.87} \\
   &    SSIM$\uparrow$ &0.449&0.466&0.521&0.480&\underline{0.600}&0.460&0.465& \textbf{0.758}\\
   &LPIPS$\downarrow$  &0.318&0.306&0.285 &0.296&\underline{0.292}&0.318&0.323 & \textbf{0.155} \\
        & CIEDE2000 $\downarrow$ & 45.72 & 46.81 & 37.11 & 36.39 & \underline{33.79} & 36.67 & 35.15 & \textbf{28.80}\\
        \hline
    \hline
    \multirow{4}*{{\textbf{HazeRD}}}& PSNR$\uparrow$ & 15.92&15.03&15.64&15.32 &15.76&15.67&\underline{16.09}& \textbf{17.39} \\
    &    SSIM$\uparrow$ & 0.643&0.601&0.624&0.673 &0.608&\underline{0.679}&0.667& \textbf{0.706}\\
     &      LPIPS$\downarrow$ &0.228&0.284&0.235&0.231 &0.243&\underline{0.222} & 0.232 & \textbf{0.224}\\
     & CIEDE2000 $\downarrow$ & 31.21 & 27.83 & 26.88 &\underline{26.08} & 30.61 & 26.25 & 26.46 & \textbf{25.75}\\
    \hline
        \hline
   \multirow{3}*{{\textbf{Real World}}}&  $r_{nve}$$\uparrow$& 1.393  & {1.418} & 0.822 & 0.704&\underline {1.664} & 1.429& 0.600& \textbf{1.767}\\
      & $m_{ng}$ $\uparrow$& 1.570 & \underline{1.699} & 1.437 & 1.116& {1.587} & 1.303 & 1.123 &\textbf{2.165}\\
       &    $r_{sp}$ $\downarrow$ & 0.0003 & 0.0195 & 0.0267& 0.0010 & \underline{0.0001} & 0.0023& {0.0010} &  \textbf{0.0000} \\
        \hline
\end{tabular}\\
\begin{scriptsize}
{~~~~~~~ $^{\rm \textcolor{myred}{[a]}}$ The \textbf{bold} results indicate the best performances, and the second-best are \underline{underlined}. $\uparrow$: The larger the better. $\downarrow$: The smaller the better. } \vspace{-.06in}
\end{scriptsize}
\label{tab.result_hd}
\end{table*}

\section{Experiment}\label{s.experiment}

\subsection{Experiment Setup}\label{ss.setup}

\subsubsection{Dataset}\label{sss.dataset}
Our dehazing model is trained on outdoor datasets, including RESIDE dataset \cite{li2018benchmarking}, NH-HAZE dataset \cite{nhhaze} and NHR dataset \cite{zhang2020nighttime} for homogeneous haze, non-homogeneous haze and nighttime haze removal, respectively. For RESIDE dataset, we randomly select 41240 samples from OTS \cite{li2018benchmarking} (outdoor training subset in RESIDE) for training and 500 samples from SOTS \cite{li2018benchmarking} (synthetic objective testing subset in RESIDE) for testing. For the NH-HAZE dataset, we synthetic 9800 samples augmented from 50 original high-resolution samples by randomly cropping, flipping, and rotating; and the 41$\sim$45-th samples are used for qualitative evaluations. The NHR dataset contains 17900 samples; we select the last 475 for testing and others for training. We also collect 372 and 132 real-world daytime and nighttime hazy images to evaluate the performance of our model.

\subsubsection{Implemental Details}\label{sss.details}

Our dehazing model is implemented with PyTorch library and trained on one NVIDIA GeForce RTX 3090 GPU.  Our model is trained for 32 epochs on the RESIDE dataset, 20 epochs on the NH-HAZE dataset and 5 epochs on the NHR dataset. We adopt Adam \cite{kingma2014adam} optimizer with the initial learning rate is $10^{-4}$  for both CNN and vision transformer backbones. We use the layer normalization \cite{ba2016layer} in the vision transformer blocks and Activation Normalization \cite{kingma2018glow} in CNN layers.
The parameters of the loss functions are set as ($\lambda_{{R}}$,$ \lambda_{{S}}$, $\lambda_{{D}}$) = $(1, 1, 1.5)$. For the HyLoG-ViT block, we set $N_l=8$ and the patch size as $2\times 2$ in the local vision transformer path; and in the global vision transformer path, we set $N_g=4$ and the patch size as $4\times 4$. The code of our model is available at \url{https://github.com/phoenixtreesky7/CFEN-ViT-Dehazing}.

\subsubsection{Metrics}
We use the peak signal-to-noise ratio (PSNR), structural similarity index measure (SSIM) \cite{wang2004image}, learned perceptual image patch similarity (LPIPS) \cite{zhang2018unreasonable} and CIEDE2000 \cite{sharma2005ciede2000} to evaluate the dehazing methods on the synthetic datasets. LPIPS measures the `perceptual distance' of the two compared images using deep features, which coincides well with human perceptual similarity judgments \cite{gonzalez2018image}.
We use the CIEDE2000 to measure the color difference between the dehazed image and its corresponding clear image. We set the parameters [$K_L$, $K_1$, $K_2$] in CIEDE2000 are [2, 0.045, 0.015] in the experiments.
Another metric we used is the bind contrast enhancement assessment \cite{hautiere2008blind} which contains three indicators, that are: 1) the rate of newly visible edges ($r_{nve}$) to evaluate the ability of edge restoration; 2) the geometric mean of the normalized gradient ($m_{ng}$) to measure the quality of the image contrast; 3) the rate of saturated pixels ($r_{sp}$) to measure the degree of over-saturation \cite{hautiere2008blind}. These three indicators are used to evaluate dehazing models on real-world datasets.

\subsection{Comparisons with State-of-the-art Methods}\label{ss.sota}

\subsubsection{Homogeneous Dehazing}


To demonstrate the effectiveness of our dehazing model on homogeneous image dehazing, we firstly evaluate it on the synthetic datasets, including SOTS (outdoor) \cite{li2018benchmarking}, O-HAZE \cite{ancuti2018ohaze}, and HazeRD \cite{zhang2017hazerd}.  We compare our method with the prior-based and learning-based models, including DCP \cite{he2011single}, HLP \cite{berman2018single}, MSCNN \cite{ren2016single}, GDNet \cite{liu2019griddehazenet}, EPDN \cite{qu2019enhanced}, MSBDN \cite{dong2020multi}, and FFA \cite{qin2020ffa}. 

It can be found from Fig.\ref{fig.result_hd} (\emph{i}), over-saturation phenomenon emerges at the sky regions in the prior-based methods of DCP and HLP, indicating that those priors may be invalid under such conditions.
The deep learning-based approaches can provide visually pleasant dehazed results on the SOTS dataset \cite{li2018benchmarking}. The metrics of PSNR, SSIM and LPIPS also demonstrate that GDNet and MSBDN obtain the top two scores on the SOTS dataset.
However, those methods exhibit relatively poor performances on the other two challenging datasets, either leading to color distortion results (such as EPDN) or leaving haze at distant scenes (such as GDNet, MSBDN and FFA), as illustrated in Fig.\ref{fig.result_hd} (\emph{ii}) and (\emph{iii}).
Our model obtains the best PSNR, SSIM, LPIPS and CIEDE2000 on the O-HAZE and HazeRD datasets.
\begin{figure*}[t!]
\centering
\includegraphics[width=.8\textwidth]{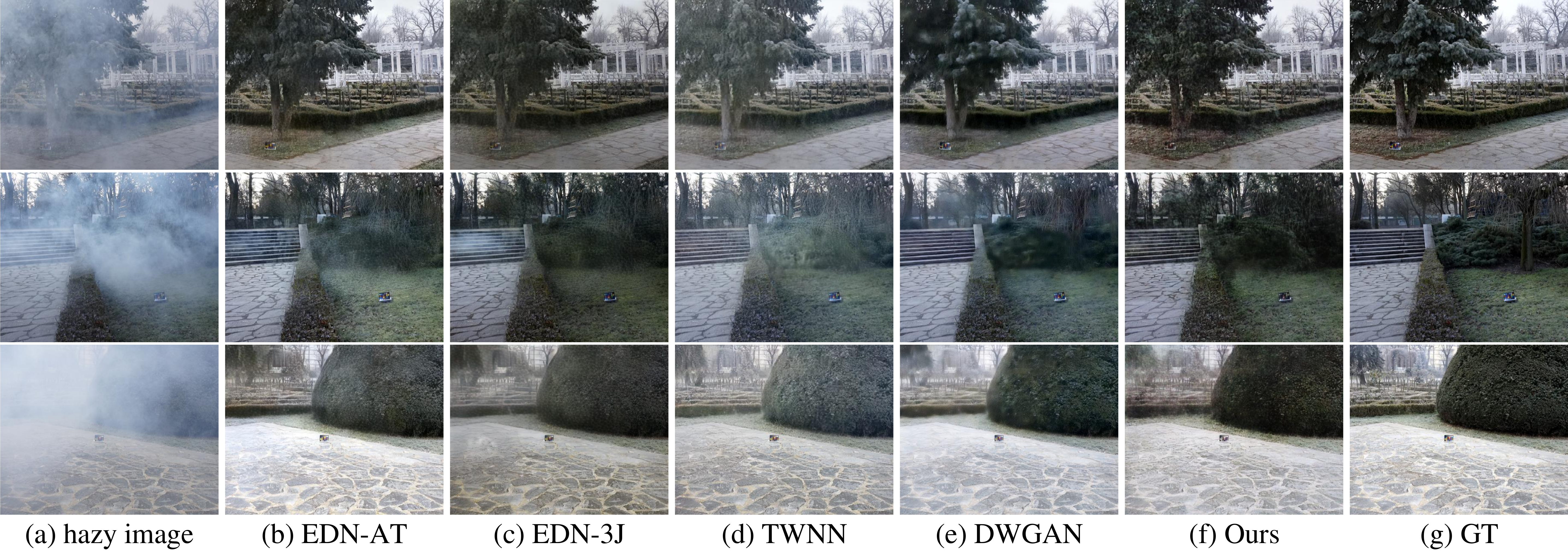} 
\caption{Dehazing results of the \emph{Non-Homogeneous} hazy images (on the NH-HAZE dataset). (a): the hazy images. (b)-(f): the dehazing results of EDN-AT \cite{edn2020cvprw}, EDN-3J\cite{edn2020cvprw}, TWNN \cite{twobranch2021cvprw}, DWGAN \cite{dwgan2021cvprw} and Ours, respectively. (g): the ground truth.}\vspace{-.08in}
\label{fig.result_nhd}
\end{figure*}

We further evaluate our method on real-world datasets. The visual comparisons are shown in Fig.\ref{fig.result_hdrw}.
The prior-based methods, i.e. DCP and HLP, tend to yield over-enhanced visual artifacts, especially in the sky regions. Although the CNNs-based methods mitigate the over-estimation artifacts successfully, some of them would produce insufficient dehazing in distant scenes. For example, the dehazed images of MSBDN and FFA appeared almost indistinguishable from the original hazy ones. The reason might lie in the overfitting on the training datasets. As obviously shown in Fig.\ref{fig.result_hdrw} (b),  our dehazing model produces plausible visual dehazing results with vivid color and fine details.
The clear superiority of our model on the real-world dataset is attributed to the complementary feature enhanced framework. The experimental results also demonstrate that our model learns the color- and texture-wise complementary features effectively.


\subsubsection{Non-Homogeneous Dehazing}

We then evaluate our method on the non-homogeneous dehazing dataset NH-HAZE. Our model is compared with several latest non-homogeneous dehazing models, including EDN-AT \cite{edn2020cvprw}, EDN-3J\cite{edn2020cvprw}, TWNN \cite{twobranch2021cvprw} and DWGAN \cite{dwgan2021cvprw}.
Fig.\ref{fig.result_nhd} reveals the visual comparisons. As we can see, all of the methods remove the non-homogeneous haze successfully. However, the EDN-3J fails to remove haze from the stone steps; the other compared methods seem to fail to restore accurate color at the bushes and lawns (see the second row in Fig.\ref{fig.result_nhd}). While our model successfully avoids these issues due to the reflectance subnetwork learning accurate color information, achieving more accurate dehazing results on the NH-HAZE dataset than other methods. The quantitative results are shown in Table \ref{tab.result_nhd} also reveal that our model obtains the best results, surpassing the second best methods by a considerable margin, surpassing the second best methods 2.13 dB on PSNR, 0.005 on SSIM, 0.025 on LPIPS, and 0.48 on CIEDE2000, respectively. The reason may lie in the proposed HyLoG-ViT, which has a stronger ability in modelling long-range context information than the vanilla CNN.
\begin{table}[t!]
\centering
\caption{Non-homogeneous dehazing results on the NH-HAZE dataset.}
\resizebox{1\columnwidth}{!}{
\begin{tabular}{c|r|rrrrrrr}
\hline
Dataset & Metric  & {\textbf{EDN-AT}} & {\textbf{EDN-3J}} & {\textbf{TBNN}} & {\textbf{DWGAN}} & {\textbf{Our}}   \\
\hline
   \multirow{4}*{{\makecell[c]{\textbf{NH-} \\ \textbf{HAZE}}}} &  PSNR$\uparrow$ & 18.92&17.78&17.78&\underline{22.13}&\textbf{24.26 }\\
   & SSIM$\uparrow$ & 0.778 & 0.735 & 0.706& \underline{0.800} & \textbf{0.805}\\
     &   LPIPS$\downarrow$ & 0.235 & 0.271 & 0.272& \underline{0.224} & \textbf{0.199} \\
       & CIEDE2000 $\downarrow$ &  23.80 & 24.29 & 26.50 & \underline{23.62} & \textbf{23.14}\\
    \hline
\end{tabular}}\vspace{-.12in}
\label{tab.result_nhd}
\end{table}

\begin{figure}[t!]
\centering
\includegraphics[width=1\columnwidth]{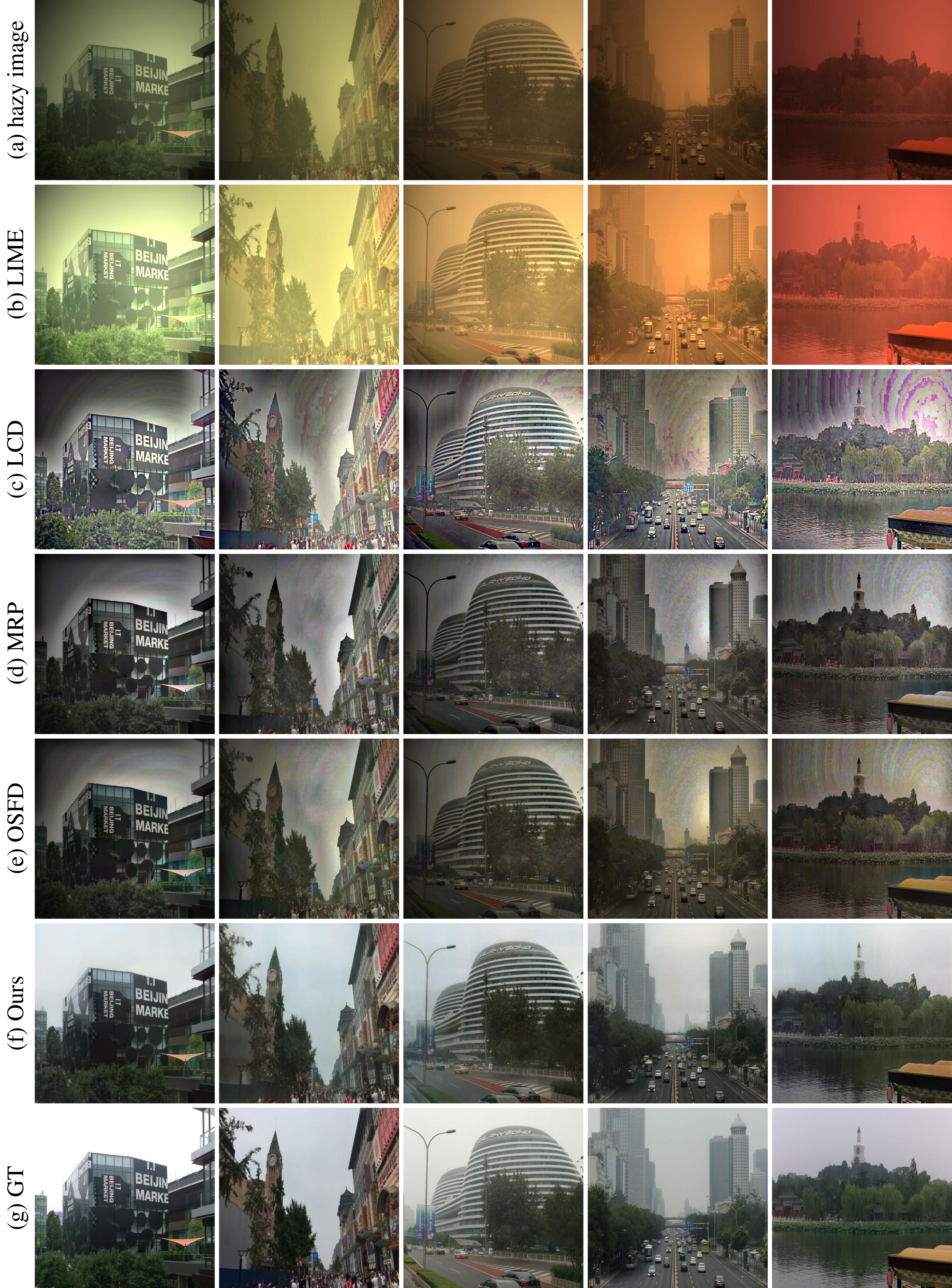}
\caption{Dehazing results of the \emph{Synthetic Nighttime} hazy images (on NHR dataset). (a): the hazy images. (b)-(f): the dehazing results of LIME \cite{guo2016lime}, LCD \cite{zhang2014nighttime}, MRP \cite{zhang2017fast}, OSFD \cite{zhang2020nighttime} and Ours, respectively. (g): the ground truth.}\vspace{-.08in}
\label{fig.result_nths}
\end{figure}
\begin{figure}[t!]
\centering
\includegraphics[width=1\columnwidth]{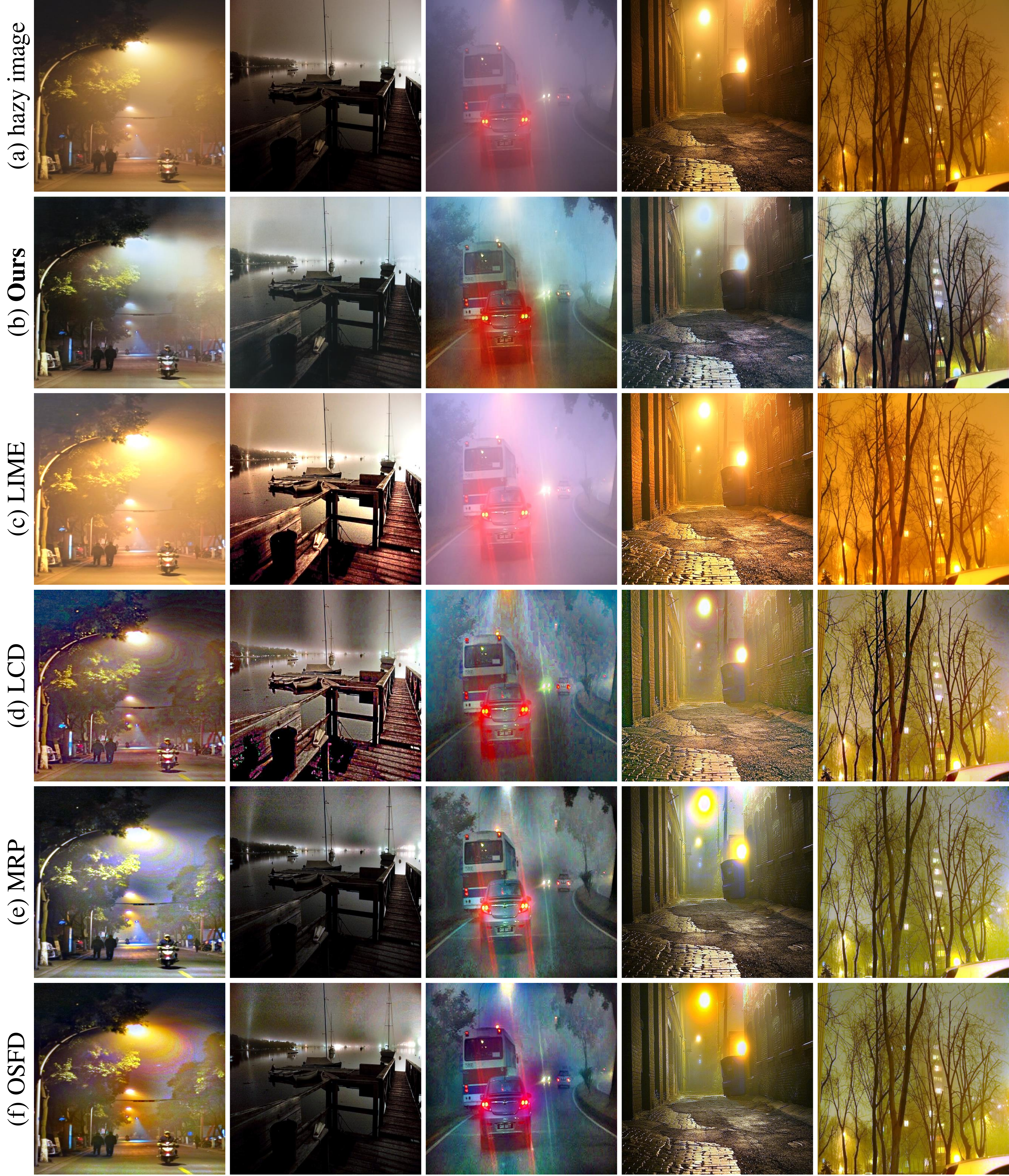}
\caption{Dehazing results of the\emph{ Real-World Nighttime} hazy images. (a): the hazy images. (b): Our dehazing results. (c)-(f): the dehazing results of LIME \cite{guo2016lime}, LCD \cite{zhang2014nighttime}, MRP \cite{zhang2017fast} and OSFD \cite{zhang2020nighttime}, respectively. }\vspace{-.08in}
\label{fig.result_nthrw}
\end{figure}

\subsubsection{Nighttime Dehazing}

We further evaluate our dehzaing model on the nighttime dehazing, comparing with a low-light image enhancement and several latest nighttime dehazing models, i.e. the LIME \cite{guo2016lime}, LCD \cite{zhang2014nighttime}, MRP \cite{zhang2017fast} and OSFD \cite{zhang2020nighttime}.
The visual results tested on the synthetic NHR dataset \cite{zhang2020nighttime} and real-world dataset are illustrated in Fig.\ref{fig.result_nths} and Fig.\ref{fig.result_nthrw}, respectively.
The low-light image enhancement method LIME only improves the luminance of the nighttime image while failing to remove the color shift and haze. Our method and the other nighttime dehazing models can improve the luminance, correct color shift and remove haze simultaneously. However, all of LCD, MRP and OSFD yield unnatural results at the sky regions in Fig.\ref{fig.result_nths} and bright regions around the street lamps in Fig.\ref{fig.result_nthrw} with over-saturation artifacts. While our method produces pleasingly smooth results in such regions, which looks more natural.
The qualitative results as shown in Table \ref{tab.result_nthd} also demonstrate that our model can perform surprisingly well on nighttime dehazing, outperforming the state-of-the-art MRP and OSFD methods with respect to PSNR, SSIM, LPIPS and CIEDE2000. Especially, our CIEDE2000 is significantly lower than the MRP with 14.54, indicating that the color-wise complementary features learned from the reflectance subnetwork play a beneficial role in preserving accurate color.

\begin{table}[t!]
\centering
\caption{Nighttime dehazing results on the NHR and Real-World datasets.}
\small
\resizebox{1\columnwidth}{!}{
\begin{tabular}{c|r|rrrrr}
\hline
Dataset & Metric & {\textbf{LIME}}  &\textbf{LCD}&\textbf{MRP}& \textbf{OSFD} & \textbf{Ours}  \\
\hline
   \multirow{4}*{{\textbf{NHR}}}& PSNR$\uparrow$ & 13.93 & 13.31&15.067 & \underline{19.88} & \textbf{22.94} \\
&    SSIM$\uparrow$&\underline{ 0.767} &0.608 &0.694 &0.700 & \textbf{0.814} \\
&    LPIPS$\downarrow$& 0.470 & 0.439 & 0.387&\underline{0.371} & \textbf{0.207} \\
      & CIEDE2000 $\downarrow$ & 43.36 & 35.02 & \underline{32.79} & 33.35 & \textbf{18.25}\\
    \hline
    \hline
   \multirow{3}*{{\makecell[c]{\textbf{Real-} \\ \textbf{World}}}}& $r_{nve}$ $\uparrow$ & 0. 144&\underline{ 0.190}&0.169& 0.149 & \textbf{0.219} \\
&    $m_{ng}$ $\uparrow$ & 5.049 &\underline{5.589} & 5.406& 4.906 & \textbf{5.846} \\
&    $r_{sp}$ $\downarrow$&\underline{ 0.0104}& 0.0136 & 0.0190& 0.0163 & \textbf{0.0000} \\
    \hline
\end{tabular}}\vspace{-.12in}
\label{tab.result_nthd}
\end{table}
\subsection{Discussion}
Due to the stronger ability in modelling long-range context information, the network structure with transformer blocks is naturally good at learning the \emph{spatially variant features} \cite{wang2021uformer}.  Our experiments verify this conclusion.
The samples in the SOTS are synthetic by setting the atmospheric light and scattering coefficient as global constants. While in the O-HAZE and NH-HAZE datasets, hazy images are captured in natural scenes where the haze is generated by professional haze machines \cite{ancuti2018ohaze}. Under foggy and hazy conditions, the atmospheric light and scattering coefficient will no longer be globally constant in natural scenes.
Therefore, as shown in Table \ref{tab.result_hd} and \ref{tab.result_nhd}, our method can surpass the SOTAs by a considerable margin on the O-HAZE and NH-HAZE, performing relatively better than on the homogeneous haze SOTS dataset.

\subsection{Ablation Study}\label{ss.ablation}
For quick experiments, all of the models are trained for 10 epochs and evaluated on 300 samples randomly selected from the SOTS \cite{li2018benchmarking} dataset in our ablation studies. 

\subsubsection{Evaluations on the Framework}\label{sss.ablation_cf}
We conduct the ablation studies to demonstrate the contribution of the joint model.
1) \textbf{w/o-RS}: our dehazing network without any complementary features, removing the decoders $D_R$ and $D_S$; 2) \textbf{w/o-S}: our dehazing model removing the decoder $D_S$; 3) \textbf{w/o-R}: our dehazing model removing the decoder $D_R$.
Comparing models of w/o-RS, w/o-S, and w/o-R, the latter two models obtain higher PSNR and SSIM than the former, indicating that the joint learning mechanism with complementary features of reflectance or shading indeed boosts the network's dehazing performance. Furthermore, when both complementary features are leveraged in our model, it achieves the best performance, demonstrating the effectiveness of the proposed framework.

\subsubsection{Evaluations on the CFSM}\label{sss.ablation_nl}
For checking the contribution of the CFSM, we conduct the experiment where the model \textbf{w/o-CFSM} refers to the dehazing model that with-out CFSM yet merge the complementary features $d_{R}^z$, $d_{S}^z$ and features $d_{D}^z$ by summation.
The dehazing results are shown in the second row in Tabel \ref{tab.ablation}.
As we can find, without the CFSM, the dehazing results reduce 0.671 dB on PSNR and 0.0029 on SSIM, indicating the importance and benefits of using the CFSM in the dehazing network.

\subsubsection{Evaluations on the HyLoG-ViT}\label{sss.ablation_vit}
We further evaluate the effectiveness of the proposed HyLoG-ViT model for image dehazing by involving the following different configurations, where the HyLoG-ViT block is replaced by:
1) two basic ResNet \cite{he2016deep} blocks (\textbf{CNN}); 2) the basic ViT block (\textbf{ViT}); 3) the local ViT path (\textbf{L-ViT}); 4) the global ViT path (\textbf{G-ViT}); 5) the sequential stacked local and global ViT paths (\textbf{LoG-ViT}).
The model CNN achieves better dehazing performance than the models of ViT and G-ViT. One of the reasons is that Transformers-based models lack some of the inductive biases inherent to model CNN, such as translation equivariance and locality, and therefore do not generalize well when trained on insufficient data. Models of LoG-ViT and HyLoG-ViT obtain the top two PSNR and SSIM, indicating that combining local and global interactions are more effective than other models. Comparing them, we can find that our parallel hybrid scheme is better than the sequentially stacked one.

\begin{table}[t]
\centering
\caption{Ablation studies on the Framework and CFSM.}
\resizebox{1\columnwidth}{!}{
\begin{tabular}{c|rrr|r|r}
\hline
   Metric      & \textbf{w/o-RS} & \textbf{w/o-S} & \textbf{w/o-R} & \textbf{w/o-CFSM} & \textbf{Our}  \\
\hline
    PSNR$\uparrow$ & 27.248 & 28.740 &27.895& 28.911 &\textbf{29.582} \\
    SSIM$\uparrow$ & 0.9419 & 0.9507 & 0.9599 & 0.9586 & \textbf{0.9617}\\
    \hline
\end{tabular}}
\label{tab.ablation}
\end{table}

\begin{table}[t]
\centering
\caption{Ablation study on the HyLoG-ViT.}
\resizebox{1\columnwidth}{!}{
\begin{tabular}{c|rrrrr|r}
\hline
  Metric   & \textbf{CNN} & \textbf{ViT} & \textbf{L-ViT} & \textbf{G-ViT} & \textbf{LoG-ViT} & \textbf{Our}    \\
     \hline
    PSNR$\uparrow$ & 28.020 & 27.727 & 28.375 & 27.365 & 28.913&\textbf{29.582} \\
    SSIM$\uparrow$ & 0.9529 & 0.9404 & 0.9582 & 0.9467 &  0.9591& \textbf{0.9617}  \\
    \hline
\end{tabular}}\vspace{-.12in}
\label{tab.ablation}
\end{table}

\subsubsection{Position Encoding}
In this experiment, we will show that, without the position encoding/embedding, our model can also achieve high performance on image dehazing. We also verify that the 3$\times$3 convolution layer that fuses the hybrid features from local and global ViT paths is useful. Hence, we compared the following models: 1) \textbf{PE}: our HyLoG adds the learnable position encodings as used in \cite{vit, chen2021pre}; 2) \textbf{ADD}: the hybrid features are fused by element-wise summation. The quantitative results are shown in Table \ref{tab.pe}. As we can find, there are no significant performance gains to the SSIM when adding the position encoding. The PE model gets even lower the PSNR than Our model.
However, the ADD model performs worse than the other two models with respect to the PSNR and SSIM, demonstrating that the 3$\times$3 convolution layer for hybrid features' fusion is vital in the HyLoG-ViT.

\begin{table}[t]
\centering
\caption{Ablation study on the position encoding.}
\small
\begin{tabular}{c|rr|r}
\hline
Metric & \textbf{PE} & \textbf{ADD} &\textbf{Our}  \\
\hline
    PSNR$\uparrow$ & 29.49 & 28.34 & \textbf{29.58} \\
    SSIM$\uparrow$& \textbf{0.963} & 0.955 & {0.962} \\
    \hline
\end{tabular}\vspace{-.12in}
\label{tab.pe}
\end{table}

\section{Conclution}
This paper aims to mitigate the issues of CNNs-based dehazing networks by introducing a new complementary feature enhanced framework and a hybrid local and global vision transformer. To these ends, we first propose a new dehazing framework, which jointly learns the intrinsic image decomposition and dehazing. The reflectance and shading prediction tasks encourage the networks to learn more useful complementary features for image dehazing task. We propose a complementary features selection module to effectively aggregate those complementary features to enhance the useful complementary features while weakening the irrelevant ones.
Then, we introduce vision transformers into the dehazing task. We design a new hybrid local-global vision transformer (HyLoG-ViT) which is more computationally efficient than the standard ViT, as it can capture both local and global dependencies. We conduct extensive experiments on homogeneous, non-homogeneous, and nighttime dehazing tasks to evaluate our method. Qualitative and quantitative results reveal that our method achieves comparable or even better performance than CNNs-based dehazing methods, demonstrating the effectiveness of the proposed framework and the HyLoG-ViT. We strongly believe that the complementary feature enhanced framework can be deeply explored and achieve outstanding performances on other image restoration tasks, such as image deraining, low-light image enhancement and old photo restoration.


{\small
\bibliographystyle{IEEEtran}
\bibliography{tnnls_dehazing}
}
\section*{Acknowledgment}
. This work is supported by grants from National Natural Science Foundation of China (No.61922006, No.62132002) and CAAI-Huawei MindSpore Open Fund.
\appendices

\ifCLASSOPTIONcaptionsoff
  \newpage
\fi

%
%
%

%
\vspace{-.395in}
\begin{IEEEbiography}[{\includegraphics[width=1in,height=1.25in,clip,keepaspectratio]{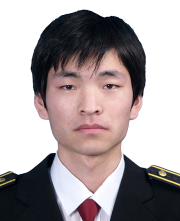}}]{Dong Zhao}
received the M.S. degree in Key Laboratory of Complex System Intelligent Control and Decision, Department of Automation, Beijing Institute of Technology, Beijing, China, in 2016, and the Ph.D. degree from National Astronomical Observatories, Chinese Academy of Sciences , Beijing, China, in 2020. His research interests include computer vision and image processing.
\end{IEEEbiography}\vspace{-.395in}

\begin{IEEEbiography}[{\includegraphics[width=1in,height=1.25in,clip,keepaspectratio]{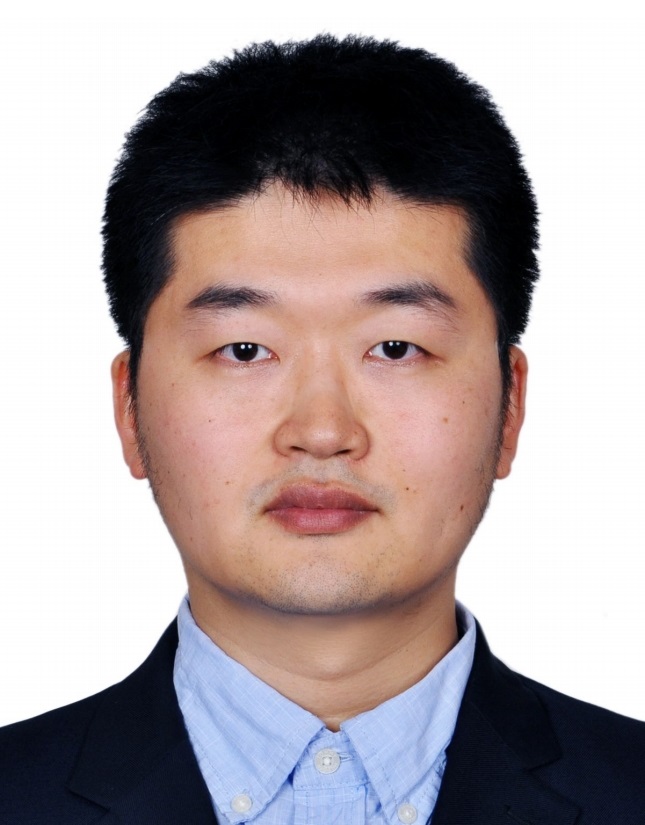}}]{Jia Li}
(M’12-SM’15) received the B.E. degree from Tsinghua University in 2005 and the Ph.D. degree
from the Institute of Computing Technology, Chinese
Academy of Sciences, in 2011. He is currently a
Full Professor with the School of Computer Science
and Engineering, Beihang University, Beijing, China.
Before he joined Beihang University in Jun. 2014,
he used to conduct research in Nanyang Technological University, Peking University and Shanda
Innovations. He is the author or coauthor of over 70
technical articles in refereed journals and conferences
such as TPAMI, IJCV, TIP, CVPR and ICCV. His research interests include
computer vision and multimedia big data, especially the understanding and
generation of visual contents. He is supported by the Research Funds for
Excellent Young Researchers from National Nature Science Foundation of
China since 2019. He was also selected into the Beijing Nova Program
(2017) and ever received the Second-grade Science Award of Chinese Institute
of Electronics (2018), two Excellent Doctoral Thesis Award from Chinese
Academy of Sciences (2012) and the Beijing Municipal Education Commission
(2012), and the First-Grade Science-Technology Progress Award from Ministry
of Education, China (2010). He is a senior member of IEEE, CIE and CCF.
More information can be found at \url{http://cvteam.net}.
\end{IEEEbiography}\vspace{-.395in}
\begin{IEEEbiography}[{\includegraphics[width=1in,height=1.25in,clip,keepaspectratio]{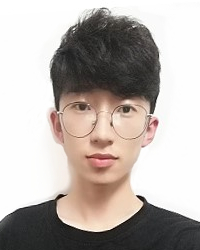}}]{Hongyu Li}
 is currently pursuing the master degree with the State Key Laboratory of Virtual Reality Technology and System, School of Computer Science and Engineering, Beihang University. He received the B.E. degree from Beihang University in Jul. 2020. His research interests include computer vision and image processing.
\end{IEEEbiography}\vspace{-.395in}

\begin{IEEEbiography}[{\includegraphics[width=1in,height=1.25in,clip,keepaspectratio]{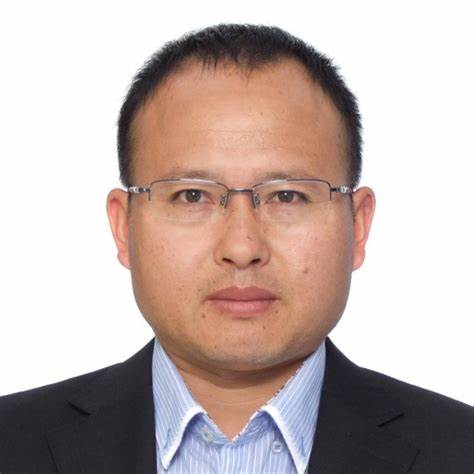}}]{Long Xu}
(M’12) received his M.S. degree in applied
mathematics from Xidian University, Xi’an, China,
in 2002, and the Ph.D. degree from the Institute
of Computing Technology, Chinese Academy of
Sciences, Beijing, China. He was a Postdoc with the
Department of Computer Science, City University
of Hong Kong, the Department of Electronic Engineering, Chinese University of Hong Kong, from
July Aug. 2009 to Dec. 2012. From Jan. 2013 to
March 2014, he was a Postdoc with the School
of Computer Engineering, Nanyang Technological
University, Singapore. Currently, he is with the Key Laboratory of Solar
Activity, National Astronomical Observatories, Chinese Academy of Sciences.
His research interests include image/video processing, solar radio astronomy,
wavelet, machine learning, and computer vision.
He was selected into the 100-Talents Plan, Chinese Academy of Sciences,
2014.
\end{IEEEbiography}
\end{document}